\documentclass[letterpaper]{article} 
\usepackage{aaai2026}  
\usepackage{times}  
\usepackage{helvet}  
\usepackage{courier}  
\usepackage[hyphens]{url}  
\usepackage{graphicx} 
\usepackage{natbib}  
\usepackage{caption} 
\usepackage{algorithm}
\usepackage{algorithmic}

\usepackage{newfloat}
\usepackage{listings}
\DeclareCaptionStyle{ruled}{labelfont=normalfont,labelsep=colon,strut=off} 
\floatstyle{ruled}
\newfloat{listing}{tb}{lst}{}
\floatname{listing}{Listing}
\title{TGDD: Trajectory Guided Dataset Distillation with Balanced Distribution}
\author {
    Fengli Ran\textsuperscript{\rm 1,2},
    Xiao Pu\textsuperscript{\rm 1},
    Bo Liu\textsuperscript{\rm 1}\realthanks,
    Xiuli Bi\textsuperscript{\rm 1}\thanksstar,
    Bin Xiao\textsuperscript{\rm 1,3},
}
\affiliations {
    \textsuperscript{\rm 1}Chongqing University of Posts and Telecommunications\\
    \textsuperscript{\rm 2}Chongqing Polytechnic University of Electronic Technology\\
    \textsuperscript{\rm 3}Jinan Inspur Data Technology Co., Ltd.\\
    fengliran@hotmail.com, \{puxiao, boliu, bixl, xiaobin\}@cqupt.edu.cn
}

\usepackage{bibentry}

\usepackage{amsmath}
\usepackage{amssymb}
\usepackage{subfigure}
\usepackage{cite}
\usepackage[table,xcdraw]{xcolor}
\usepackage{booktabs, multirow}
\usepackage{makecell}   

\renewcommand{\surd}{$\checkmark$}  

\newcommand{\pmnum}[2]{%
  #1\,\scalebox{0.75}{$\pm$}\,\scalebox{0.75}{#2}%
}

\makeatletter
\newcommand{\realthanks}{%
  \thanks{Corresponding Author}%
}

\newcommand{\thanksstar}{%
  \footnotemark[\value{footnote}]%
}
\makeatother

\begin{document}

\maketitle

\begin{abstract}
Dataset distillation compresses large datasets into compact synthetic ones to reduce storage and computational costs. Among various approaches, distribution matching (DM)-based methods have attracted attention for their high efficiency. However, they often overlook the evolution of feature representations during training, which limits the expressiveness of synthetic data and weakens downstream performance.
To address this issue, we propose Trajectory Guided Dataset Distillation (TGDD), which reformulates distribution matching as a dynamic alignment process along the model’s training trajectory. At each training stage, TGDD captures evolving semantics by aligning the feature distribution between the synthetic and original dataset. Meanwhile, it introduces a distribution constraint regularization to reduce class overlap. This design helps synthetic data preserve both semantic diversity and representativeness, improving performance in downstream tasks. Without additional optimization overhead, TGDD achieves a favorable balance between performance and efficiency. Experiments on ten datasets demonstrate that TGDD achieves state-of-the-art performance, notably a 5.0\% accuracy gain on high-resolution benchmarks. 

\end{abstract}

\begin{links}
    \link{Code}{https://github.com/FlyFinley/TGDD}
\end{links}

\section{Introduction}
In the deep learning era, the exponential expansion of dataset sizes has brought substantial computational and storage burdens. For instance, data volumes have ballooned from ImageNet’s tens of millions \cite{krizhevsky2017imagenet} to LAION‑5B’s billions of image–text pairs \cite{schuhmann2022laion}. Reducing large datasets to smaller ones that can preserve performance under resource constraints has become a key challenge in advancing AI.

To tackle this challenge, a strategy is coreset selection ~\cite{feldman2020core,chai2023efficient}, which picks a small representative subset from the original dataset. However, this approach discards most samples, overlooks their training potential, and diminishes the information available for downstream tasks. Another strategy is dataset distillation~\cite{cazenavette2023generalizing,yang2024neural}, which directly synthesizes a compact set with higher information density to approximate the performance of the full dataset.

\begin{figure}[t!]
    \centering
    \includegraphics[width=0.95\linewidth]{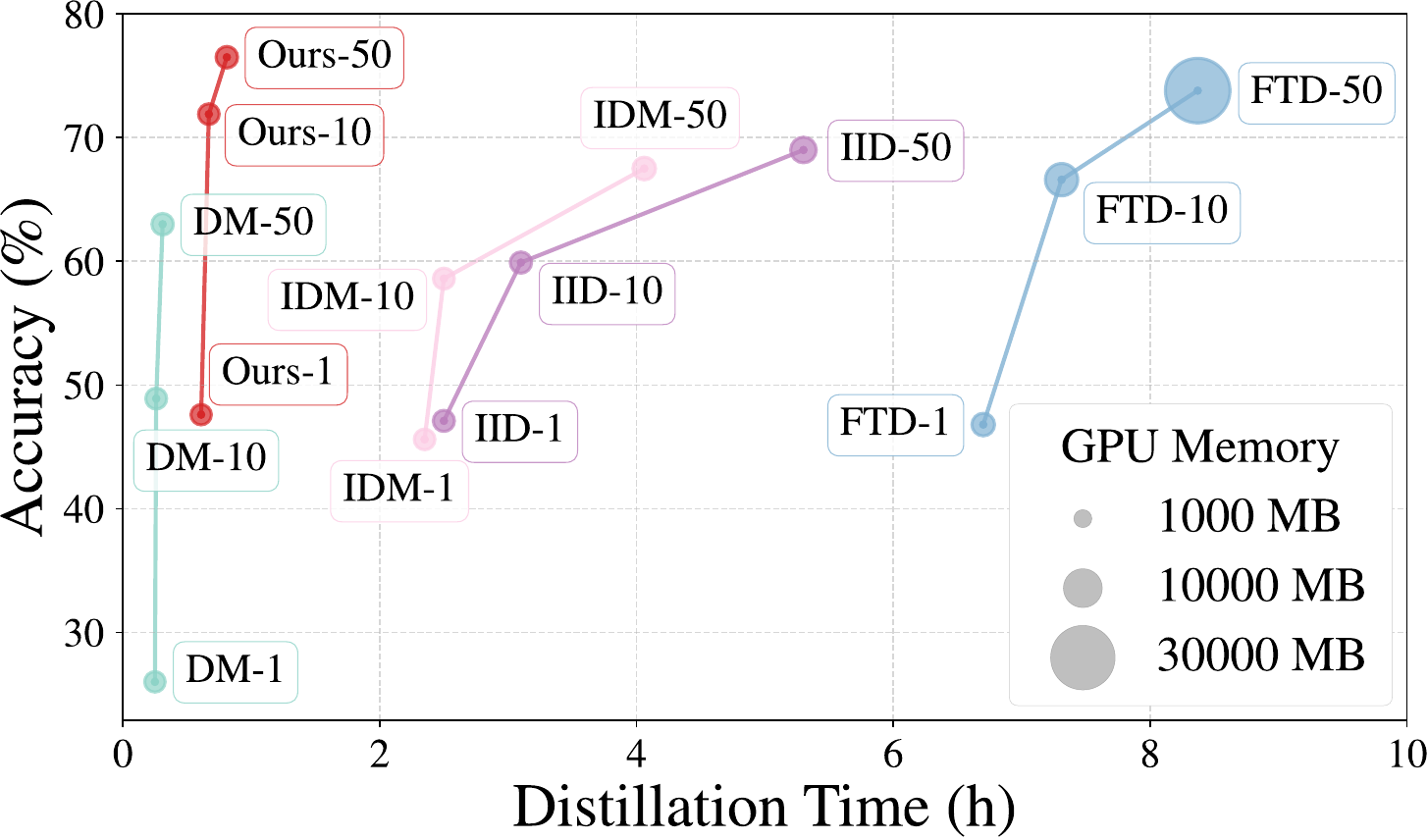}
    \caption{
    Accuracy, distillation time, and GPU memory comparison on CIFAR-10 under different IPCs. Pretraining time is included (5 trajectories for ours, 100 for FTD). Our method balances performance and cost effectively.
    } 
    \label{fig:performance_comparison}
\end{figure}

Existing dataset distillation methods generally fall into two categories: optimization-oriented (OO)-based and distribution-matching (DM)-based approaches. OO‑based techniques recreate the learning dynamics by forcing synthetic data to induce the same gradient~\cite{zhao2021dataset,kim2022dataset} or parameter~\cite{cazenavette2022dataset,cui2023scaling}  as original data throughout training. As shown in Figure \ref{fig:performance_comparison}, although methods like FTD \cite{du2023minimizing} are effective, their iterative updates between data and model incur heavy computational costs, limiting their scalability to large datasets. In contrast, DM-based methods focus on the statistical properties of the data, directly aligning synthetic and original feature distributions in the embedding space without requiring model updates, thereby dramatically accelerating the distillation process.

\begin{figure}[t!]
    \centering
    \includegraphics[width=0.95\linewidth]{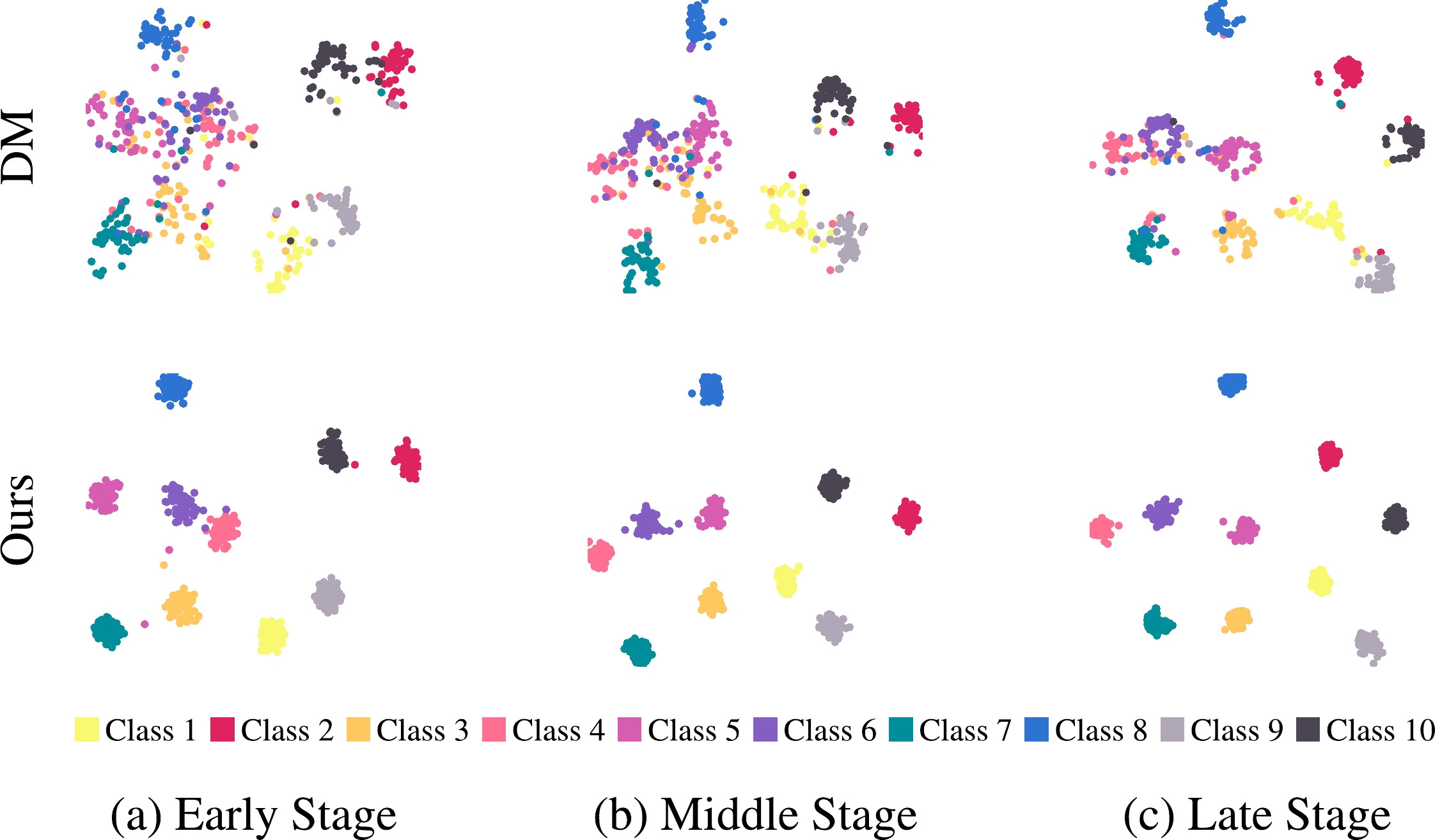}
    \caption{
    t-SNE visualization of synthetic features generated by DM and our method under IPC-50, using pretrained models on CIFAR-10 at different training stages.
    } 
    \label{fig:motivation}
\end{figure}

A critical problem for DM-based methods is effectively mapping data into the embedding space. However, most DM-based approaches adopt randomly initialized networks for feature extraction, overlooking how model representations evolve throughout training \cite{zeiler2014visualizing, rahaman2019spectral}, which yields inadequate embeddings. As shown in Figure \ref{fig:motivation}, DM’s synthetic datasets exhibit significantly different class separability across models; only highly optimized networks can distinguish categories effectively. This makes synthetic sets hard to learn and undermines downstream performance. 

To address these issues, we propose Trajectory Guided Dataset Distillation (TGDD), which recasts the static distribution matching into a dynamic balancing process along the trajectory. 
Specifically, TGDD first builds expert trajectories by saving model snapshots at different training stages. Leveraging these trajectories, TGDD then aligns the feature distributions of the synthetic and original data at every stage, enriching the semantic diversity of the synthetic dataset.
Simultaneously, TGDD applies a stage-wise distribution constraint regularization, enforcing intra-class compactness and improving representativeness of the synthetic dataset.
As shown in Figure \ref{fig:motivation}, TGDD balances the diversity and representativeness of the synthetic dataset throughout training, making it a good surrogate of the original dataset and improving its downstream performance. 
Since expert trajectories are constructed using only the original dataset, they can be pretrained and reused in different settings. TGDD achieves an favorable balance between distillation performance and computational efficiency.

In summary, our contributions are as follows:
\begin{itemize}
    \item We reformulate distribution matching as a dynamic balance process between feature alignment and distribution constraint throughout training, unifying representativeness and diversity in the synthetic distribution.
    \item We introduce Trajectory-Guided Distribution Distillation (TGDD), which leverages multiple pretrained expert trajectories to perform efficient feature alignment and stage-wise distribution constraint during training.
    \item Extensive experiments across ten datasets demonstrate that TGDD achieves state‑of‑the‑art performance, notably, a 5.0\% accuracy gain is achieved on high‑resolution benchmarks.
\end{itemize}

\section{Related Work}
\subsection{Coreset selection}
Coreset selection \cite{chai2023efficient,lee2024coreset} aims to extract a small, representative subset from a large dataset and has been widely adopted in continual learning \cite{rebuffi2017icarl,aljundi2019gradient} and active learning \cite{agarwal2020contextual,kim2022defense}. 
Early approaches rely on uniform random sampling, while later work develops more advanced criteria. Herding \cite{welling2009herding} iteratively minimizes the distance between the subset and the full‑set feature centroids. K‑Center \cite{sener2017active} refines this idea by choosing samples that minimize the dataset’s maximum covering radius. Forgetting \cite{toneva2018empirical} retains examples most prone to being forgotten during training. Despite their efficiency, these approaches irrevocably discard significant portions of the original data and thus cannot guarantee optimal downstream performance \cite{lei2023comprehensive}. 

\subsection{Dataset Distillation}

Dataset distillation compresses large datasets into smaller ones while preserving original information \cite{yu2023dataset}, benefiting tasks such as neural architecture search \cite{such2020generative}, federated learning \cite{pi2023dynafed, wang2024aggregation}, continual learning \cite{yang2023efficient}, and privacy protection \cite{dong2022privacy}. Existing methods can be broadly categorized as follows.

\textbf{Optimization-Oriented (OO)-Based Methods} cast dataset distillation as bilevel optimization, updating model parameters on synthetic data while refining that data to preserve original performance. The pioneering work DD \cite{wang2018dataset} introduces the concept of dataset distillation from a meta learning perspective. Later, DC \cite{DBLP:conf/iclr/ZhaoMB21} expands the idea by matching gradients from the original and synthetic data. DSA \cite{zhao2021dataset} further enhances performance by incorporating differentiated Siamese augmentation. 
Instead of matching gradients, MTT \cite{cazenavette2022dataset} aligns model parameters over training trajectories. 
FTD \cite{du2023minimizing} mitigates trajectory drift by reducing cumulative errors across steps. 
DATM \cite{guo2024lossless} scales trajectory matching by adapting sample difficulty to dataset size.
Despite improved performance, their bilevel optimization incurs significant computational cost.

\textbf{Distribution-Matching (DM)-Based Methods} directly align the feature distributions of synthetic and original data in latent space, bypassing bilevel optimization to improve efficiency. DM \cite{zhao2023dataset} introduces the concept by aligning class-wise centroids. Although fast, its performance is limited by oversimplified feature matching. 
DataDAM \cite{sajedi2023datadam} extends this idea by incorporating multi-layer alignment through attention mechanisms. 
IDM \cite{zhao2023improved} adopts a dynamic model queue to extract more informative representations, but incurres extra training costs. M3D \cite{zhang2024m3d} further projectes features into a Reproducing Kernel Hilbert Space for finer alignment. DANCE \cite{zhang2024dance} interpolates initial and converged models to create pseudo-intermediate feature extractors. 
Although distribution matching shortens distillation time, it ignores representation evolution and lacks inter-class compactness, leading to scattered features and poor separability. We address these gaps with trajectory-aware alignment and compactness regularization to produce highly discriminative synthetic data.

\begin{figure*}[t!]
    \centering
    \includegraphics[scale=0.642]{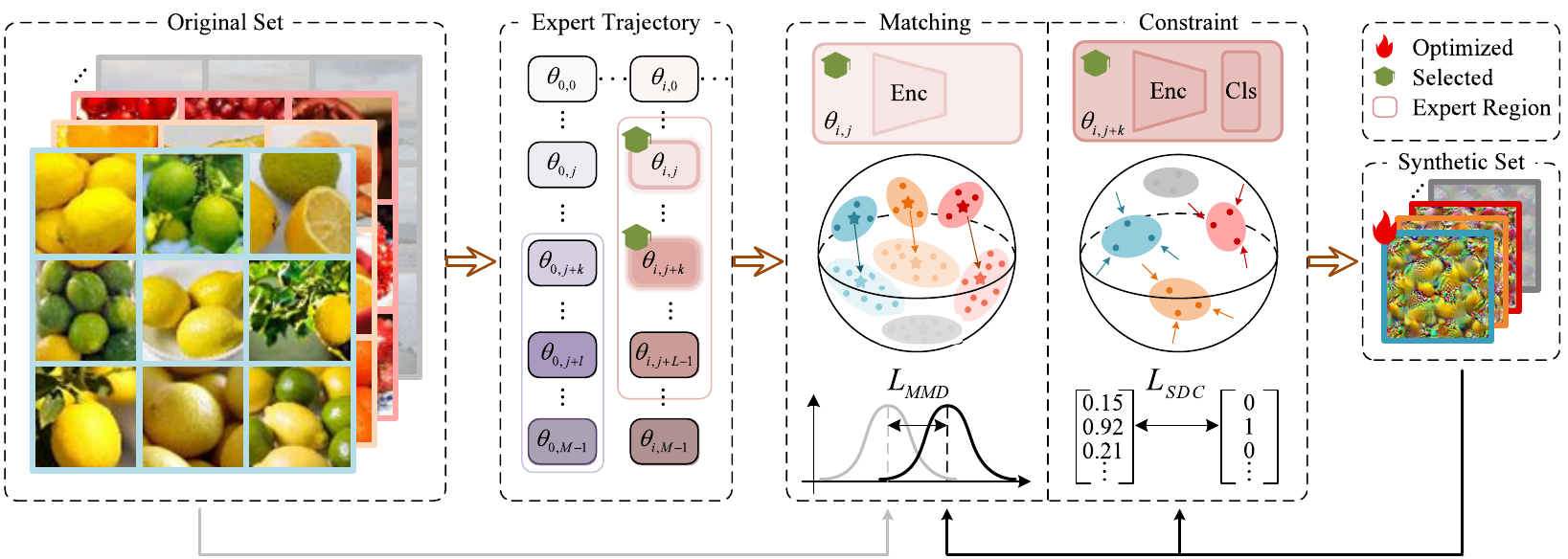}
    \caption{
    The illustration of our proposed method. First, we pretrain N expert trajectories with original dataset, each comprising M network snapshots. Then, one snapshot is randomly sampled as encoder for distribution matching between original and synthetic dataset, another snapshot in the expert region is chosen to impose distribution constraints on the synthetic dataset.
    }
    \label{fig:Method Overview}
\end{figure*}

\section{Method}
\subsection{Preliminary}
\subsubsection{Dataset distillation}
Given a large source training dataset \(T=\left \{ (x_{i},y_{i}) \right \} |_{i=1}^{\left | T \right | } \) , the objective of Dataset Distillation is to synthesize a smaller target synthetic dataset \(S=\left \{ (s_{j},s_{j}) \right \} |_{j=1}^{\left | S \right | } \) , where \(\left | S \right | \ll \left | T \right |\). When we evaluate the original data on models trained with \(T\) and \(S\) respectively, we hope to have a similar generalization performance. To be specific, let \(\left ( x_{i},y_{i} \right ) \) be the data pair sampled from the original data distribution, \(l\) be the classification loss function such as cross entropy, \(\theta ^{T}\) and \(\theta ^{S}\) be the models trained from \(T\) and \(S\) respectively, our aim is to minimize the generalization performance gap:
\begin{equation}
E_{x\sim P_{D}}[(l_{\theta ^{T}}(x) ,y)]\simeq E_{x\sim P_{D}}[(l_{\theta ^{S}}(x) ,y)]
\end{equation}
\subsubsection{Distribution Matching}
Prior methods for dataset distillation mostly rely on OO-based techniques like gradient or trajectory matching. As these methods require simultaneous optimization of the network and synthetic dataset, they suffer from high computational overhead and struggle to scale to large datasets. To improve computational efficiency, the DM-based approaches constrain the embedding features of the original and synthetic datasets through the maximum mean discrepancy (MMD) \cite{gretton2012kernel}, requiring only optimization of the synthetic dataset. We have:
\begin{equation}
S^{*}  = \arg \min_{S} \mathbb{E}_{\theta \sim P_{\theta_{0}} } \left \|\frac{1}{T }\sum_{i=1}^{\left | T \right | } \psi_{\theta}(x_{i} ) - \frac{1}{S}\sum_{j=1}^{\left | S \right | } \psi_{\theta}(s_{j})    \right \| ^{2}  \label{origin_MMD}
\end{equation}
Here \(\theta \sim P_{\theta_{0}}\), most distribution matching methods employ randomly initialized networks as feature extractors.

\subsection{Expert Trajectory Construction}
Expert trajectories comprise model snapshots captured at different epochs using original data, defining the performance upper bound for models trained on the dataset. Trajectory-matching methods leverage these trajectories as their foundation. Despite achieving strong performance, such methods incur substantial storage overhead because of preserving a large number of trajectories. For instance, MTT \cite{cazenavette2022dataset} stores 200 trajectories.

In contrast, conventional DM-based approaches typically employ randomly initialized models, overlooking the potential of expert trajectories. We demonstrate that strategic integration of expert trajectories significantly enhances distribution matching for dataset distillation. Our framework therefore incorporates expert trajectories by training N randomly initialized neural networks that converge within M epochs, thereby forming expert trajectories as:
\begin{equation}
\mathbf{P} = \left\{p_{i,j} \mid 0 \leq 
i \leq N, \ 0 \leq j \leq M\right\}.
\end{equation}
Since expert trajectories require only original data for training, we can pretrain models before distillation. This approach retains efficiency of distribution matching, significantly faster than OO-based methods that train networks during distillation. 
Moreover, our method achieves competitive performance with even one trajectory, substantially reducing storage overhead compared to trajectory matching techniques that require massive trajectory preservation.

\subsection{Stage-wise Distribution Matching}
Conventional DM-based methods typically employ randomly initialized networks as feature extractors, aligning features between synthetic and original data. However, these homogeneous initialization strategies capture only early-stage feature distributions, neglecting evolving patterns throughout the training trajectory.

Inspired by OO-based methods that perform gradient matching or trajectory matching across the whole training process, we leverage pretrained expert trajectories for feature distribution at different training stages. This approach eliminates network training during distillation for efficiency while providing diverse feature representations spanning all optimization phases.

During each distillation iteration, we randomly select an expert trajectory $\mathbf{P}_{i}$ and then sample a pretrained model $\theta_{ext}=p_{i,j}$ at arbitrary training stage. The encoder component of this model computes representations for the original dataset and the synthetic dataset. We then align these representations through distribution matching.

We formulate the goal of the stage-wise distribution matching as follows:

\begin{equation}
    L_{MMD}  = \sum_{c=1}^{C} \left \|\frac{1}{\left | B_{c}^{T}  \right |}  \sum_{i=1}^{\left | B_{c}^{T}  \right | } \psi_{\theta_{ext}}(x_{i})-\frac{1}{\left | B_{c}^{S}  \right |} \sum_{i=1}^{\left | B_{c}^{S}  \right | } \psi_{\theta_{ext}}(s_{i}) \right \| ^{2}
    \label{new_MMD}
\end{equation}
where $\theta_{ext}$ is the feature extractor network weight sampled from the expert trajectories, \(B_{c}^{T}\) and \(B_{c}^{S}\) are mini-batch pairs sampled from \(T\) and \(S\) respectively for class c.

\subsection{Stage-wise Distribution Constraint}
Conventional distribution matching methods rely exclusively on Maximum Mean Discrepancy (MMD) for feature alignment. However, MMD constrains only mean of feature distributions, providing insufficient regularization. This leads to overly dispersed synthetic data distributions and severe class boundary confusion.

Effective distribution constraints are therefore crucial for enhancing inter-class discriminability in synthetic datasets. By fully exploiting expert trajectories, we introduce a distribution constraint regularization that operates on expert regions. Using different experts in training iterations, the method achieves ensemble-like performance without incurring additional training cost. 

Specifically, at each step of the distillation process, once the feature extractor $\theta_{ext} = p_{i,j}$ is determined, we construct an expert region consisting of $L$ pretrained expert networks $P_{er} = \{p_{i,j},p_{i,j+1},...,p_{i,j+L-1}\}$. The expert model $\theta_{exp}=p_{i,j+k}$ is then randomly selected from the region and employed to impose the distributional constraint. We calculate the stage-wise distribution constraint loss as:

\begin{equation}
\begin{aligned}
L_{SDC}=\frac{1}{B_{c}^{S}}\sum_{c=1}^{C}  \sum_{i=1}^{\left |B_{c}^{S}   \right | } l(\phi_{exp}(s_{i}),y_{i} )
\label{expert_loss}
\end{aligned}
\end{equation}
where $\theta_{exp}$ is the expert network weight sampled from the expert region, \(B_{c}^{S}\) are mini-batch pairs sampled from \(S\) for class c.

\subsection{Training Algorithm}
The overall loss function is described in equation \ref{total_loss}
\begin{equation}
    L_{overall} =L_{MMD}+\alpha L_{SDC} \label{total_loss}
\end{equation}
where \(\mathcal{L}_{\text{MMD}}\) is the distribution matching loss, \(\mathcal{L}_{\text{SDC}}\) is the distribution constraint loss and \(\alpha\) is the regularization coefficient. Both feature extraction and constraint networks are sampled from the expert trajectories. 
Algorithm \ref{overall_alg} details our proposed method.

\begin{algorithm}[!ht]
    \renewcommand{\algorithmicrequire}{\textbf{Input:}}
    \renewcommand{\algorithmicensure}{\textbf{Output:}}
    \caption{Trajectory Guided Dataset Distillation with Balanced Distribution}
    \label{overall_alg}
    \begin{algorithmic}[1] 
        \REQUIRE original dataset \(D ^T\), regularization coefficient \(\alpha\), iteration nums \(Iter\), expert region distance \(L\), synthetic dataset learning rate \(\eta\)
        \STATE train N expert trajectories, each consists of M snapshots, \(\mathbf{P} = \left\{p_{i,j} \mid 0 \leq i \leq N, \ 0 \leq j \leq M\right\}\)
        \FOR {\(i=1,2,\cdots,Iter\)}
            \STATE Initialize synthetic dataset \(D ^S\) with random selected data from \(D ^T\)
            \STATE Sample an expert trajectory \(\mathbf{P}_{i}\ \)
            \STATE Sample a feature extract network \(p_{i,j}\) from \(\mathbf{P}_{i}\)
            \STATE Calculate the MMD loss with Equation \ref{new_MMD}
            \STATE Sample an expert network \(p_{i,j+k}\) from \(p_{i,j}\) to \(p_{i,j+L-1}\) 
            \STATE Calculate the SDC loss with Equation \ref{expert_loss}
            \STATE Calculate the total loss with Equation \ref{total_loss} and update \(D ^S\) with \(D ^S=D ^S-\eta \nabla_{\!D ^S} L_{overall}\)
        \ENDFOR
        
        \ENSURE synthetic dataset \(D_{syn}\); 
    \end{algorithmic}
\end{algorithm}

\section{Experiments}

\setlength{\tabcolsep}{1mm}
\begin{table*}[]
\centering
\small
\begin{tabular}{lcccccccccccc}
\toprule
Datasets & \multicolumn{3}{c}{SVHN} & \multicolumn{3}{c}{CIFAR-10} & \multicolumn{3}{c}{CIFAR-100} & \multicolumn{3}{c}{TinyImageNet} \\ 
\cmidrule(lr){2-4} \cmidrule(lr){5-7} \cmidrule(lr){8-10} \cmidrule(l){11-13}
IPC & 1 & 10 & 50 & 1 & 10 & 50 & 1 & 10 & 50 & 1 & 10 & 50 \\
Ratio (\%) & 0.02 & 0.14 & 0.7 & 0.02 & 0.2 & 1 & 0.02 & 0.2 & 1 & 0.2 & 2 & 10 \\ 
\cmidrule(lr){2-4} \cmidrule(lr){5-7} \cmidrule(lr){8-10} \cmidrule(l){11-13}
Whole & \multicolumn{3}{c}{\pmnum{95.4}{0.1}} & \multicolumn{3}{c}{\pmnum{84.8}{0.1}} & \multicolumn{3}{c}{\pmnum{56.2}{0.3}} & \multicolumn{3}{c}{\pmnum{37.6}{0.4}} \\ \midrule
Random & \pmnum{14.4}{0.2} & \pmnum{26.0}{1.2} & \pmnum{43.4}{1.0} & \pmnum{14.4}{0.2} & \pmnum{26.0}{1.2} & \pmnum{43.4}{1.0} & \pmnum{4.2}{0.3} & \pmnum{14.6}{0.5} & \pmnum{30.0}{0.4} & \pmnum{1.4}{0.1 } & \pmnum{5.0}{0.2} & \pmnum{15.0}{0.4} \\
Herding & \pmnum{21.5}{1.2} & \pmnum{31.6}{0.7} & \pmnum{40.4}{0.6} & \pmnum{21.5}{1.2} & \pmnum{31.6}{0.7} & \pmnum{40.4}{0.6} & \pmnum{8.4}{0.3} & \pmnum{17.3}{0.3} & \pmnum{33.7}{0.5} & \pmnum{2.8}{0.2} & \pmnum{6.3}{0.2} & \pmnum{16.7}{0.3} \\
Forgetting & \multicolumn{1}{l}{\pmnum{12.1}{5.6}} & \multicolumn{1}{l}{\pmnum{16.8}{1.2}} & \multicolumn{1}{l}{\pmnum{27.2}{1.5}} & \pmnum{13.5}{1.2} & \pmnum{23.3}{1.0} & \pmnum{23.3}{1.1} & \pmnum{4.5}{0.2} & \pmnum{15.1}{0.3} & \pmnum{30.5}{0.3} & \pmnum{1.6}{0.1} & \pmnum{5.1}{0.2} & \pmnum{15.0}{0.3} \\ \midrule
DC & \multicolumn{1}{l}{\pmnum{31.2}{1.4 }} & \multicolumn{1}{l}{\pmnum{76.1}{0.6}} & \multicolumn{1}{l}{\pmnum{82.3}{0.3}} & \pmnum{28.3}{0.5} & \pmnum{44.9}{0.5} & \pmnum{53.9}{0.5} & \pmnum{12.8}{0.3} & \pmnum{25.2}{0.3} & - & - & - & - \\
DSA & \multicolumn{1}{l}{\pmnum{27.5}{1.4}} & \multicolumn{1}{l}{\pmnum{79.2}{0.5}} & \multicolumn{1}{l}{\pmnum{84.4}{0.4}} & \pmnum{28.8}{0.7} & \pmnum{52.1}{0.5} & \pmnum{60.6}{0.5} & \pmnum{13.9}{0.3} & \pmnum{32.3}{0.3} & \pmnum{42.8}{0.4} & - & - & - \\
MTT & - & - & - & \pmnum{46.3}{0.8} & \pmnum{65.3}{0.7} & \pmnum{71.6}{0.2} & \pmnum{24.3}{0.3} & \pmnum{40.1}{0.4} & \pmnum{47.7}{0.2} & \pmnum{8.8}{0.3} & \pmnum{23.2}{0.2} & \pmnum{28.0}{0.3} \\
FTD & - & - & - & \pmnum{46.8}{0.3} & \pmnum{66.6}{0.3} & \pmnum{73.8}{0.2} & \pmnum{25.2}{0.2 } & \pmnum{43.4}{0.3} & \pmnum{50.7}{0.3} & \pmnum{10.4}{0.3} & \pmnum{24.5}{0.2} & - \\ \midrule
CAFE & \multicolumn{1}{l}{\pmnum{42.6}{3.3}} & \multicolumn{1}{l}{\pmnum{75.9}{0.6}} & \multicolumn{1}{l}{\pmnum{81.3}{0.3}} & \pmnum{30.3}{1.1} & \pmnum{46.3}{0.6} & \pmnum{55.5}{0.6} & \pmnum{12.9}{0.3} & \pmnum{27.8}{0.3} & \pmnum{37.9}{0.3} & -  & -  & - \\
DM & - & - & - & \pmnum{26.0}{0.8} & \pmnum{48.9}{0.6} & \pmnum{63}{0.4} & \pmnum{11.4}{0.3} & \pmnum{29.7}{0.3} & \pmnum{43.6}{0.4} & \pmnum{3.9}{±0.2} & \pmnum{12.9}{0.4} & \pmnum{24.1}{0.3} \\
IDM & - & - & - & \pmnum{45.6}{0.7} & \pmnum{58.6}{0.1} & \pmnum{67.5}{0.1} & \pmnum{20.1}{0.3} & \pmnum{45.1}{0.1} & \pmnum{50}{0.2} & \pmnum{10.1}{0.2} & \pmnum{21.9}{0.2} & \pmnum{27.7}{0.3} \\
M3D & \multicolumn{1}{l}{\textbf{\pmnum{62.8}{0.5}}} & \multicolumn{1}{l}{\pmnum{85.0}{0.1}} & \multicolumn{1}{l}{\pmnum{86.2}{0.3}} & \pmnum{45.3}{0.3 } & \pmnum{63.5}{0.2} & \pmnum{69.9}{0.5} & \pmnum{26.2}{0.3} & \pmnum{42.4}{0.2} & \pmnum{50.9}{0.7} & - & - & - \\
DSDM & \multicolumn{1}{l}{\pmnum{60.2}{0.2}} & \multicolumn{1}{l}{\pmnum{85.4}{0.3}} & \multicolumn{1}{l}{\pmnum{91.3}{0.2}} & \pmnum{45.0}{0.4} & \pmnum{66.5}{0.3} & \pmnum{75.8}{0.3} & \pmnum{19.5}{0.2} & \pmnum{46.2}{0.3 } & \pmnum{54.0}{0.2} & - & - & - \\
DANCE & - & - & - & \pmnum{47.1}{0.2} & \pmnum{70.8}{0.2} & \pmnum{76.1}{0.1} & \pmnum{27.9}{0.2} & \pmnum{49.8}{0.1} & \pmnum{52.8}{0.1} & \pmnum{11.6}{0.2 } & \pmnum{26.4}{0.3} & \pmnum{28.9}{0.4} \\ 
\rowcolor[HTML]{EFEFEF} \textbf{Ours} & \multicolumn{1}{l}{\pmnum{59.0}{0.7}} & \multicolumn{1}{l}{\textbf{\pmnum{88.2}{0.3}}} & \multicolumn{1}{l}{\textbf{\pmnum{92.8}{0.3}}} & \textbf{\pmnum{47.6}{0.2}} & \textbf{\pmnum{71.9}{0.3}} & \textbf{\pmnum{76.5}{0.2}} & \textbf{\pmnum{28.5}{0.2}} & \textbf{\pmnum{51.3}{0.2}} & \textbf{\pmnum{54.6}{0.3}} & \textbf{\pmnum{13.6}{0.1}} & \textbf{\pmnum{29.3}{0.3}} & \textbf{\pmnum{30.9}{0.4}} \\ \bottomrule

\end{tabular}
\caption{Comparison with previous coreset selection and dataset distillation methods on low-resolution and medium-resolution datasets. IPC: images per class. Ratio (\%): the ratio of synthetic images to the original dataset. Whole: the accuracy of the model trained with the whole training set. }
\label{table_cifar}
\end{table*}

\setlength{\tabcolsep}{1mm}
\begin{table*}[t!]
\centering
\small
\begin{tabular}{lcccccccccccc}
\toprule
 Datasets & \multicolumn{2}{c}{ImageNette} & \multicolumn{2}{c}{ImageWoof} & \multicolumn{2}{c}{ImageFruit} & \multicolumn{2}{c}{ImageMeow} & \multicolumn{2}{c}{ImageSquawk} & \multicolumn{2}{c}{ImageYellow} \\ 
 \cmidrule(lr){2-3} \cmidrule(lr){4-5} \cmidrule(lr){6-7} \cmidrule(l){8-9} \cmidrule(l){10-11} \cmidrule(l){12-13}
IPC & 1 & 10 & 1 & 10 & 1 & 10 & 1 & 10 & 1 & 10 & 1 & 10 \\
Ratio(\%) & 0.105 & 1.05 & 0.11 & 1.1 & 0.077 & 0.077 & 0.077 & 0.077 & 0.077 & 0.077 & 0.077 & 0.077  \\ 
 \cmidrule(lr){2-3} \cmidrule(lr){4-5} \cmidrule(lr){6-7} \cmidrule(l){8-9} \cmidrule(l){10-11} \cmidrule(l){12-13}
Whole & \multicolumn{2}{c}{\pmnum{87.4}{1.0}} & \multicolumn{2}{c}{\pmnum{67.0}{1.3}} & \multicolumn{2}{c}{\pmnum{63.9}{2.0}} & \multicolumn{2}{c}{\pmnum{66.7}{1.1}} & \multicolumn{2}{c}{\pmnum{87.5}{0.3}} & \multicolumn{2}{c}{\pmnum{84.4}{0.6}} \\ \midrule
Random & \pmnum{23.5}{4.8} & \pmnum{47.7}{2.4} & \pmnum{14.2}{0.9} & \pmnum{27.0}{1.9} & \pmnum{13.2}{0.8} & \pmnum{21.4}{1.2} & \pmnum{13.8}{0.6} & \pmnum{29.0}{1.1} & \pmnum{21.8}{0.5} & \pmnum{40.2}{0.4} & \pmnum{20.4}{0.6} & \pmnum{37.4}{0.5} \\
MTT & \pmnum{47.7}{0.9} & \pmnum{63.0}{1.3} & \pmnum{28.6}{0.8} & \pmnum{35.8}{1.8} & \pmnum{26.6}{0.8} & \pmnum{40.3}{1.3} & \pmnum{30.7}{1.6} & \pmnum{40.4}{2.2} & \pmnum{39.4}{1.5} & \pmnum{52.3}{1.0} & \pmnum{45.2}{0.8} & \pmnum{60.0}{1.5} \\
FTD & \pmnum{52.2}{1.0} & \pmnum{67.7}{0.7} & \pmnum{30.1}{1.0} & \pmnum{38.8}{1.4} & \pmnum{29.1}{0.9} & \pmnum{44.9}{1.5} & \pmnum{33.8}{1.5} & \pmnum{43.3}{0.6} & - & - & - & - \\
DM & \pmnum{32.8}{0.5} & \pmnum{58.1}{0.3 } & \pmnum{21.1}{1.2} & \pmnum{31.4}{0.5} & - & - & - & - & \pmnum{31.2}{0.7} & \pmnum{50.4}{1.2} & - & - \\
DANCE & \pmnum{57.2}{0.5} & \pmnum{80.2}{0.7} & \pmnum{30.6}{0.3} & \pmnum{57.8}{1.1} & \pmnum{30.6}{0.8} & \pmnum{52.8}{0.7} & \pmnum{39.4}{0.8} & \pmnum{60.4}{1.1} & \pmnum{52.0}{0.5} & \pmnum{77.2}{0.3} & \pmnum{51.8}{1.1} & \textbf{\pmnum{78.8}{0.7}} \\ 
\rowcolor[HTML]{EFEFEF} \textbf{Ours} & \textbf{\pmnum{61.8}{0.7} }& \textbf{\pmnum{82}{0.5}} & \textbf{\pmnum{34.6}{0.3}} & \textbf{\pmnum{58.4}{0.7}} & \textbf{\pmnum{34.8}{0.5}} & \textbf{\pmnum{57.8}{0.6}} & \textbf{\pmnum{41.4}{0.5}} & \textbf{\pmnum{62.8}{0.9}} & \textbf{\pmnum{53.2}{0.6}} & \textbf{\pmnum{78}{0.3}} &\textbf{ \pmnum{55.8}{0.7}} & \pmnum{76.6}{0.8} \\ \bottomrule
\end{tabular}
\caption{Comparison with previous coreset selection and dataset distillation methods on ImageNet Subset}
\label{table_imagenet}
\end{table*}

\subsection{Experimental Setup}
\textbf{Datasets.}
We assessed dataset distillation across multiple datasets, including the low-resolution SVHN \cite{netzer2011reading}, CIFAR-10 and CIFAR-100 (32×32) \cite{krizhevsky2009learning}, the medium-resolution Tiny ImageNet (64×64) \cite{le2015tiny}, and high-resolution ImageNet subsets (128×128)\cite{deng2009imagenet}, namely ImageNette, ImageWoof, ImageFruit, ImageMeow, ImageSquawk, and ImageYellow.

\subsubsection{Networks.}
Following previous work, we adopted ConvNet \cite{gidaris2018dynamic} for dataset distillation. For SVHN, CIFAR-10 and CIFAR-100, a 3-layer ConvNet was used. Each layer had a 128 kernel 3×3 convolutional kernel, instance normalization \cite{ulyanov2016instance}, ReLU activation \cite{nair2010rectified}, and a 3×3 average pooling layer with a stride of 2. For Tiny ImageNet and ImageNet subsets, a 4-layer and 5-layer ConvNet was applied respectively.

\subsubsection{Implementation Details.}
During distillation, the optimizer's learning rate is set to 0.1 for ImageNet subsets and 0.01 for others, scaled by the images per class. In training, an SGD optimizer with a 0.01 learning rate, 0.9 momentum, and 0.0005 weight decay is used. Regarding the hyperparameter \(\alpha\) , we set 2.5 for 1 and 10 images per class, and 0.5 for 50 images per class. Hyperparameter L is set to 7 in all settings. We trained 5 expert trajectoies with 60 epochs for SVHN, CIFAR-10, CIFAR-100 and Tiny ImageNet, and 80 epochs for ImageNet subsets. Following previous research \cite{zhao2021dataset}, we used differential augmentation like color transformation, random crop, cutout, random flip, scale and rotate transformation. We also use multiformation parameterization as in \cite{kim2022dataset}, where the factor parameter \(\rho\) is set to 3 for ImageNet subsets and 2 for other datasets. Each experiment was repeated 5 times.

\subsection{Comparison with Previous Methods}
\subsubsection{Performance Comparison}
We compare our method with various baseline approaches across different resolutions and scenarios. For coreset methods, baselines like Random, Herding \cite{welling2009herding}, and Forgetting \cite{toneva2018empirical} are selected. Among OO-based methods, DC \cite{DBLP:conf/iclr/ZhaoMB21}, DSA \cite{zhao2021dataset}, MTT \cite{cazenavette2022dataset} and FTD \cite{du2023minimizing} are chosen for comparison. And for DM-based methods, CAFE \cite{wang2022cafe}, DM \cite{zhao2023dataset}, IDM \cite{zhao2023improved}, M3D \cite{zhang2024m3d}, DSDM \cite{li2024diversified} and DANCE \cite{zhang2024dance} are included. Table \ref{table_cifar} and Table \ref{table_imagenet} present the comparative results of our method and other baselines across multiple benchmark datasets, such as SVHN, CIFAR-10, CIFAR-100, TinyImageNet, and subsets of ImageNet, demonstrating the effectiveness of our approach.
To be Specific, in experiments on CIFAR-10, when IPC is 10 and 50, our method surpasses the classical DM algorithm by 23\% and 13.5\%, respectively, and it also outperforms OO-based algorithms such as DC and MTT. On the TinyImageNet dataset, our approach achieves remarkable accuracy of 29.3\% at IPC-10 and 30.9\% at IPC-50, both of which are higher than the current SOTA methods. Furthermore, on the ImageNet subset, our method demonstrates superior performance. For instance, a 5\% gain improvement is achived in ImageFruit at IPC-10.

\subsubsection{Cross-Architecture Evaluation}
We evaluate the performance of synthetic datasets generated by our method across different network architectures. As shown in Table \ref{table_cross_arch}, we generate synthetic datasets using a 3-layer ConvNet and evaluate them on ResNet10 \cite{he2016deep} and DenseNet121 \cite{huang2017densely}. Our method demonstrates superior generalization performance across multiple architectures at IPC-10 and IPC-50.

\begin{figure*}[t!] 
    \centering
    \subfigure[Stage-wise Feature Correlation]
    {
        \begin{minipage}[b]{0.32\textwidth} 
            \centering           
 \includegraphics[scale=0.485]{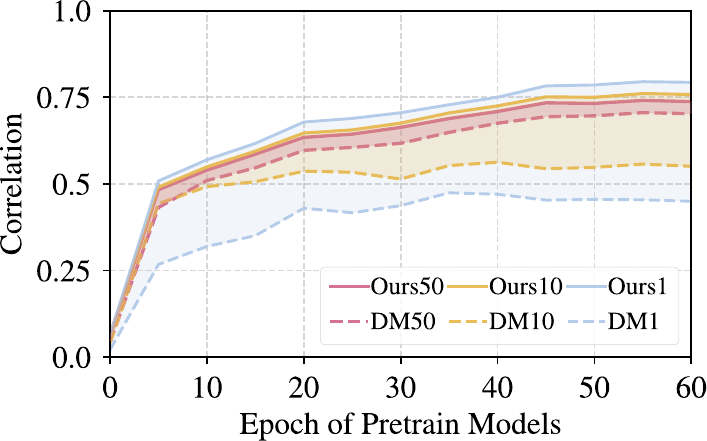}
        \end{minipage}
        \label{fig:motivation_mmd}
    }
    \subfigure[Stage-wise Snapshot Accuracy]
    {
        \begin{minipage}[b]{0.32\textwidth}
            \centering
            \includegraphics[scale=0.485]{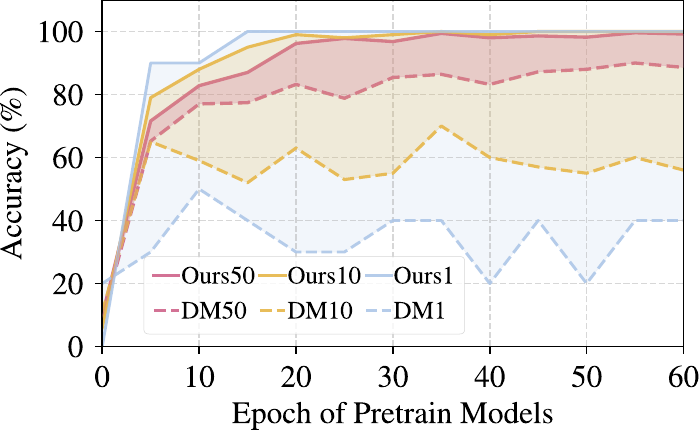}
        \end{minipage}
        \label{fig:motivation_expert}
    }
    \subfigure[Neuron activation ratio] 
    {
        \begin{minipage}[b]{0.32\textwidth}
            \centering
            \includegraphics[scale=0.485]{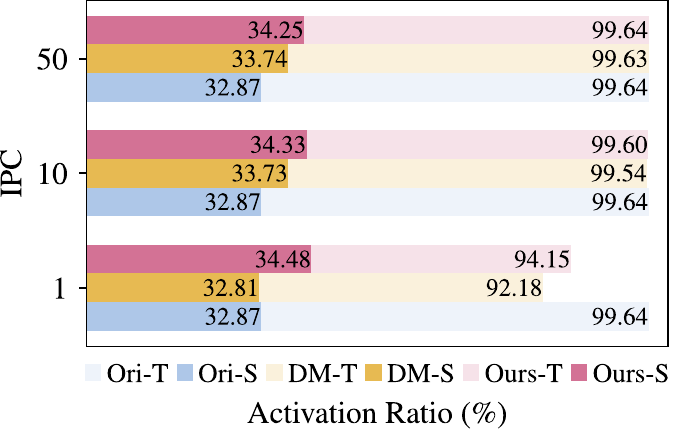}
        \end{minipage}
        \label{fig:motivation_information}
    }
    \caption{
    Effectiveness study of our method. (a) Distribution similarity via feature correlations. (b) Class separability evaluated by pretrained model accuracy across stages. (c) Information density from per-image and dataset-level neuron activation ratios.
    } 
    \label{fig:Motivation_Experiment} 
\end{figure*}

\begin{figure}[t!]
    \centering
    \includegraphics[width=0.95\linewidth]{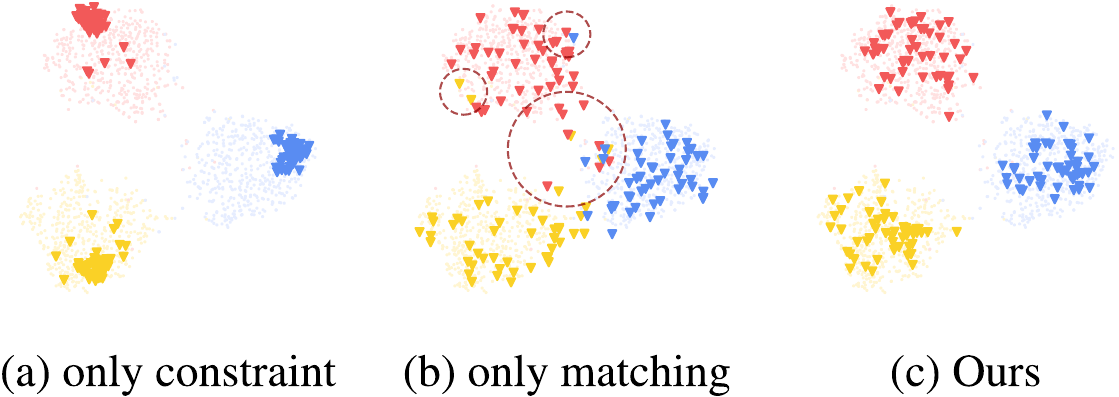}
    \caption{Distribution of original images and synthetic images on CIFAR-10 with IPC-50.}
    \label{fig:distribution_ablation}
\end{figure}

\subsection{Effectiveness of our method}
\subsubsection{Distribution Matching}
To validate the effectiveness of the proposed distribution matching method, we extract features from both the synthetic and original datasets using pretrained models at different training stages and compute their similarity.
Figure \ref{fig:motivation_mmd} illustrates that our method consistently achieves superior feature alignment across all training stages and configurations, demonstrating its improved performance in distribution matching.
In contrast, Figure \ref{fig:distribution_ablation}(a) shows that replacing the stage-wise feature extractors with randomly initialized networks leads to overly concentrated feature distributions, severely compromising data diversity.

\subsubsection{Distribution Constraint}
To assess the impact of the proposed distribution constraint, we evaluate the synthetic datasets using different pretrained models under various configurations.
Figure \ref{fig:motivation_expert} shows that our method consistently achieves superior classification accuracy across all training stages, reflecting improved class separability and reduced overlap between categories in the synthesized data.
In contrast, Figure \ref{fig:distribution_ablation}(b) highlights substantial degradation when expert models are replaced with randomly initialized ones, dispersed feature representations lead to severe entanglement and poor class distinction.

\subsubsection{Information Distillation}
 We quantify information density through the ratio of neurons activated per image and across the entire dataset, where higher activation ratio indicates greater information density. Following previous work \cite{wang2024learning}, neurons with gradients exceeding layer-wise mean values are considered activated. Figure \ref{fig:motivation_information} demonstrates that at equivalent total activation levels, our method activates more neurons per image, confirming higher information density in individual synthetic samples.

\begin{table}[b!]
\centering
\small
\setlength{\tabcolsep}{6.5pt}
\begin{tabular}{llccc}
\toprule
Method & IPC & \multicolumn{1}{l}{ConvNet-3} & \multicolumn{1}{l}{ResNet-10} & \multicolumn{1}{l}{DenseNet-121} \\ \midrule
 & 10 & \pmnum{52.1}{0.5} & \pmnum{32.9}{0.3} & \pmnum{34.5}{0.1} \\
\multirow{-2}{*}{DSA} & 50 & \pmnum{60.6}{0.5} & \pmnum{49.7}{0.4} & \pmnum{49.1}{0.2} \\ 
 & 10 & \pmnum{56.4}{0.7} & \pmnum{34.5}{0.8} & \pmnum{41.5}{0.5} \\
\multirow{-2}{*}{MTT} & 50 & \pmnum{65.9}{0.6} & \pmnum{43.2}{0.4} & \pmnum{51.9}{0.3} \\ 
 & 10 & \pmnum{48.9}{0.6} & \pmnum{42.3}{0.5} & \pmnum{39.0}{0.1} \\
\multirow{-2}{*}{DM} & 50 & \pmnum{63.0}{0.4 }&  \pmnum{58.6}{0.3} & \pmnum{57.4}{0.3} \\ 
 & 10 & \pmnum{63.5}{0.2} & \pmnum{56.7}{0.3} & \pmnum{54.6}{0.2} \\
\multirow{-2}{*}{M3D} & 50 & \pmnum{69.9}{0.5} & \pmnum{66.6}{0.3} & \pmnum{66.1}{0.4} \\ 
 & 10 & \pmnum{70.8}{0.2} & \pmnum{67.0}{0.2} & \pmnum{64.5}{0.3} \\
\multirow{-2}{*}{DANCE} & 50 & \pmnum{76.1}{0.1} & \pmnum{68.0}{0.1} & \pmnum{64.8}{0.3} \\ \midrule
\rowcolor[HTML]{EFEFEF} 
\cellcolor[HTML]{EFEFEF} & 10 & \textbf{\pmnum{71.9}{0.3}} & \textbf{\pmnum{67.7}{0.2}} & \textbf{\pmnum{68.2}{0.3}} \\
\rowcolor[HTML]{EFEFEF} 
\multirow{-2}{*}{\cellcolor[HTML]{EFEFEF}Ours} & 50 & \textbf{\pmnum{76.5}{0.2}} & \textbf{\pmnum{74.9}{0.4}} & \textbf{\pmnum{74.3}{0.2}} \\ \bottomrule
\end{tabular}
\caption{Accuracy on CIFAR-10 with different architectures. Synthetic dataset are condensed using ConvNet-3.}
\label{table_cross_arch}
\end{table}

\begin{table}[b!]
\centering
\small

\begin{tabular}{ccc cccc}

\toprule
\multirow{2.6}{*}{\makecell{Aug}} &
\multirow{2.6}{*}{\makecell{\(L_{MMD}\)}} &
\multirow{2.6}{*}{\makecell{\(L_{SDC}\)}} &
\multicolumn{2}{c}{CIFAR-10} &
\multicolumn{2}{c}{CIFAR-100} \\
\cmidrule(lr){4-5} \cmidrule(lr){6-7}
 & & & 10 & 50 & 10 & 50 \\
\midrule

- & - & - & \pmnum{55.2}{0.2}  & \pmnum{65.3}{0.3} & \pmnum{33.7}{0.2} & \pmnum{44.5}{0.3} \\
\(\surd \) & - & - & \pmnum{63.2}{0.2} & \pmnum{69.5}{0.3} & \pmnum{40.5}{0.2} & \pmnum{47.2}{0.3} \\
\(\surd \) & \(\surd \) & - & \pmnum{65.8}{0.3} & \pmnum{75.2}{0.2} & \pmnum{47.0}{0.3} & \pmnum{53.0}{0.2} \\
\(\surd \) & \(\surd \) & \(\surd \) & \cellcolor[HTML]{EFEFEF}\textbf{\pmnum{71.9}{0.3}} & \cellcolor[HTML]{EFEFEF}\textbf{\pmnum{76.5}{0.2}} & \cellcolor[HTML]{EFEFEF}\textbf{\pmnum{51.3}{0.2}} & \cellcolor[HTML]{EFEFEF}\textbf{\pmnum{54.6}{0.3}} \\ \bottomrule
\end{tabular}
\caption{Accuracy ablation study of each component on CIFAR-10/100 with IPC-10 and IPC-50.}
\label{table_ablation}
\end{table}

\begin{figure}[t!]
\centering
\subfigure[Regularization coefficient \(\alpha\)]
{
    \begin{minipage}[b]{0.47\linewidth}
        \includegraphics[scale=0.35]{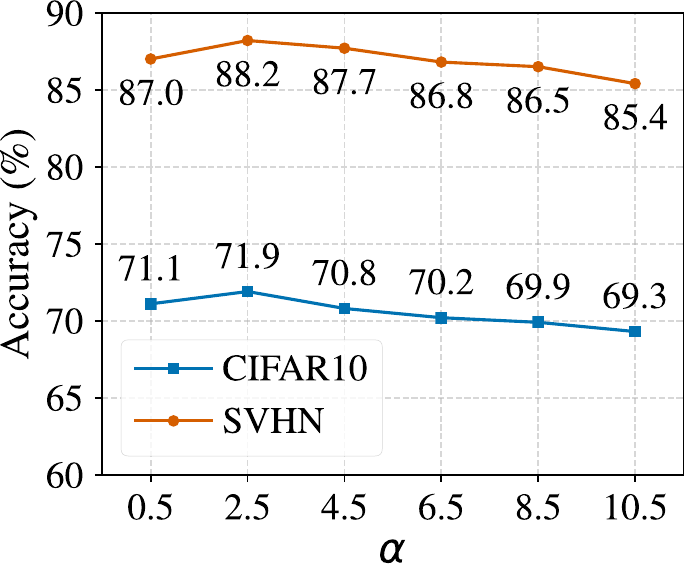}
    \end{minipage}
    \label{fig:hyper_alpha}
}
\subfigure[Length of expert area]
{
 	\begin{minipage}[b]{0.47\linewidth}
        \includegraphics[scale=0.35]{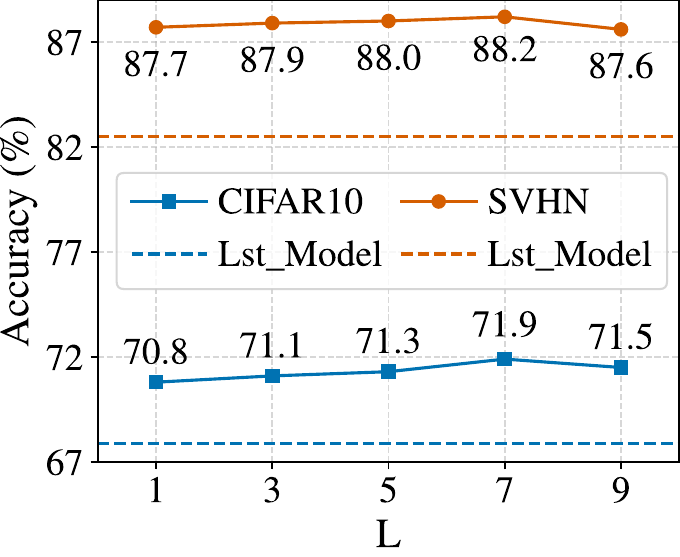}
    \end{minipage}
    \label{fig:hyper_range}
}
\subfigure[Number of trajectory]
{
    \begin{minipage}[b]{0.47\linewidth}
        \includegraphics[scale=0.35]{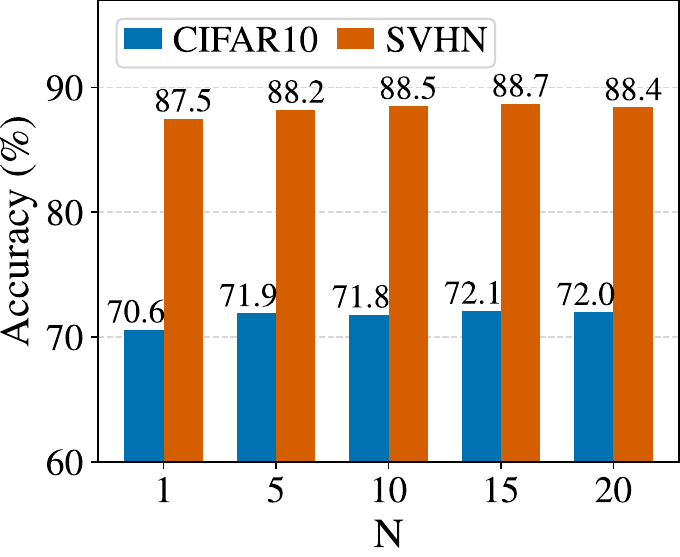}
    \end{minipage}
    \label{fig:hyper_npm}
}
\subfigure[Length of trajectory]
{
 	\begin{minipage}[b]{0.47\linewidth}
        \includegraphics[scale=0.35]{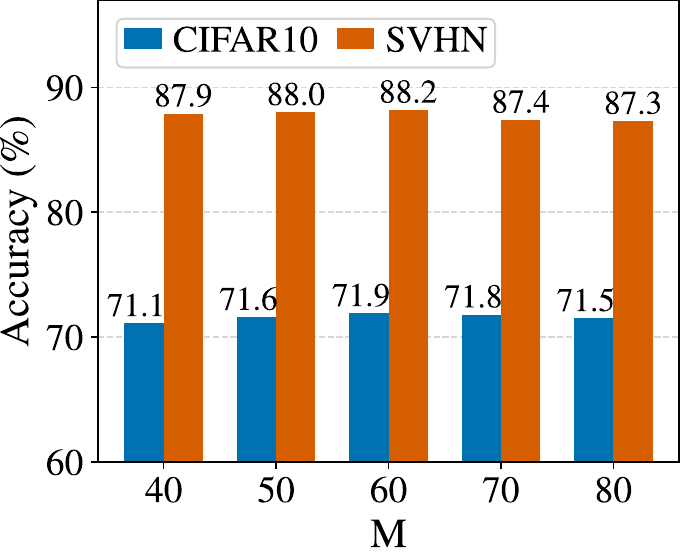}
    \end{minipage}
    \label{fig:hyper_epoch}
}
\caption{Effect of different hyperparameters with IPC-10.}
\label{fig:hyper_parameter}
\end{figure}

\subsection{Ablation Study}
\subsubsection{Effectiveness of each Component}
We evaluate the impact of the main modules of the proposed method on performance, namely multiformation augmentation, stage-wise distribution matching, and stage-wise distribution constraints.
As illustrated in Table \ref{table_ablation}, the proposed module achieves performance gains across multiple datasets at various IPC. Optimal performance is achieved when all modules are used in combination, indicating the effectiveness of the proposed method.

\subsubsection{Impact of \(\alpha\)}
The regularization coefficient \(\alpha\) reflects the importance of expert distribution constraints in the loss. The figure \ref{fig:hyper_alpha} shows how performance varies with \(\alpha\). When \(\alpha\) ranges from 0.5 to 10.5, the model's performance fluctuates within the range of 2.8\%, indicating moderate model sensitivity to this hyperparameter.

\subsubsection{Impact of expert region}
The expert region refers to a set of model snapshots within a range of length L starting from the feature extractor. During distillation, we randomly sample expert models from this region to impose distributional constraints. As shown in Figure \ref{fig:hyper_range}, using the last converged model as the baseline, our expert region design demonstrates consistent performance across various \(L\), and significantly outperforms the single-expert setting.

\begin{figure}[t!]
\centering
\subfigure[step=5]
{
    \begin{minipage}[b]{0.47\linewidth}
        \centering
        \includegraphics[scale=0.35]{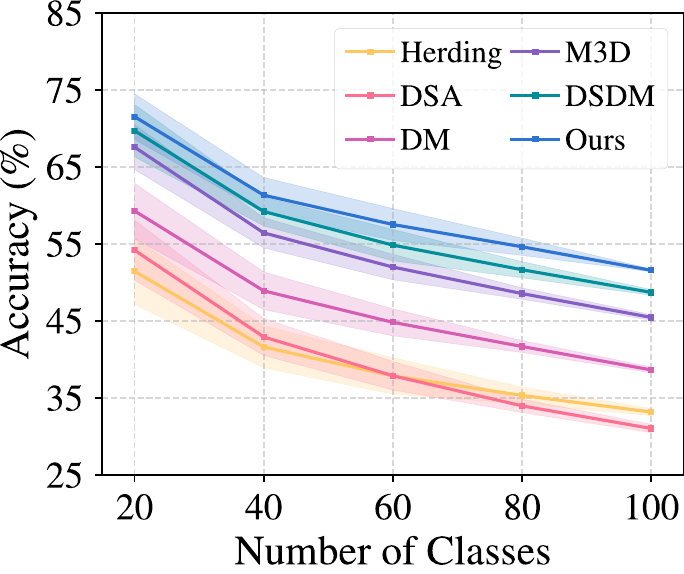}
    \end{minipage}
}
\subfigure[step=10]
{
 	\begin{minipage}[b]{0.47\linewidth}
        \centering
        \includegraphics[scale=0.35]{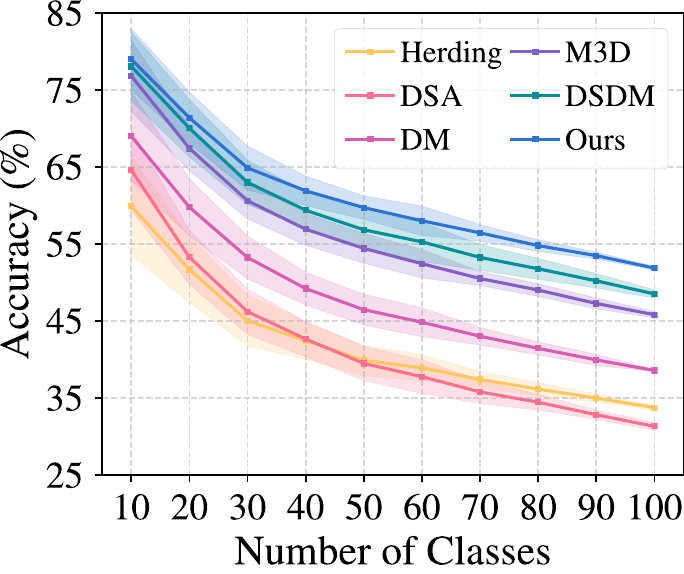}
    \end{minipage}
}
\caption{Class-incremental learning on CIFAR-100. }
\label{fig:continual_learning}
\end{figure}

\subsubsection{Impact of expert trajectory}
The expert trajectory, characterized by its number and length, affects both distribution matching and constraint. As illustrated in Figure \ref{fig:hyper_npm}, distillation performance improves slightly with an increasing number of trajectories, but gradually reaches saturation. Similarly, Figure \ref{fig:hyper_epoch} shows that extending trajectory length yields comparable trends in performance gain. To balance performance and efficiency, we refrain from excessively increasing the number or length of expert trajectories.

\subsection{Continual Learning}
Continual learning aims to train models to adapt to new tasks while preserving performance on previous ones and minimizing catastrophic forgetting \cite{prabhu2020gdumb}. Dataset distillation, by reducing the volume of original dataset, shows great potential in this area. We compare Herding, DSA, DM, DSDM, M3D and our method. Following prior studies, we keep 20 images per class, set step sizes at 5 and 10, use a 3-layer ConvNet, and run each experiment 5 times to get mean and variance values. As demonstrated in Figure \ref{fig:continual_learning}, our approach significantly outperforms baselines.

\begin{figure}[t!]
    \centering
    \includegraphics[width=0.96\linewidth]{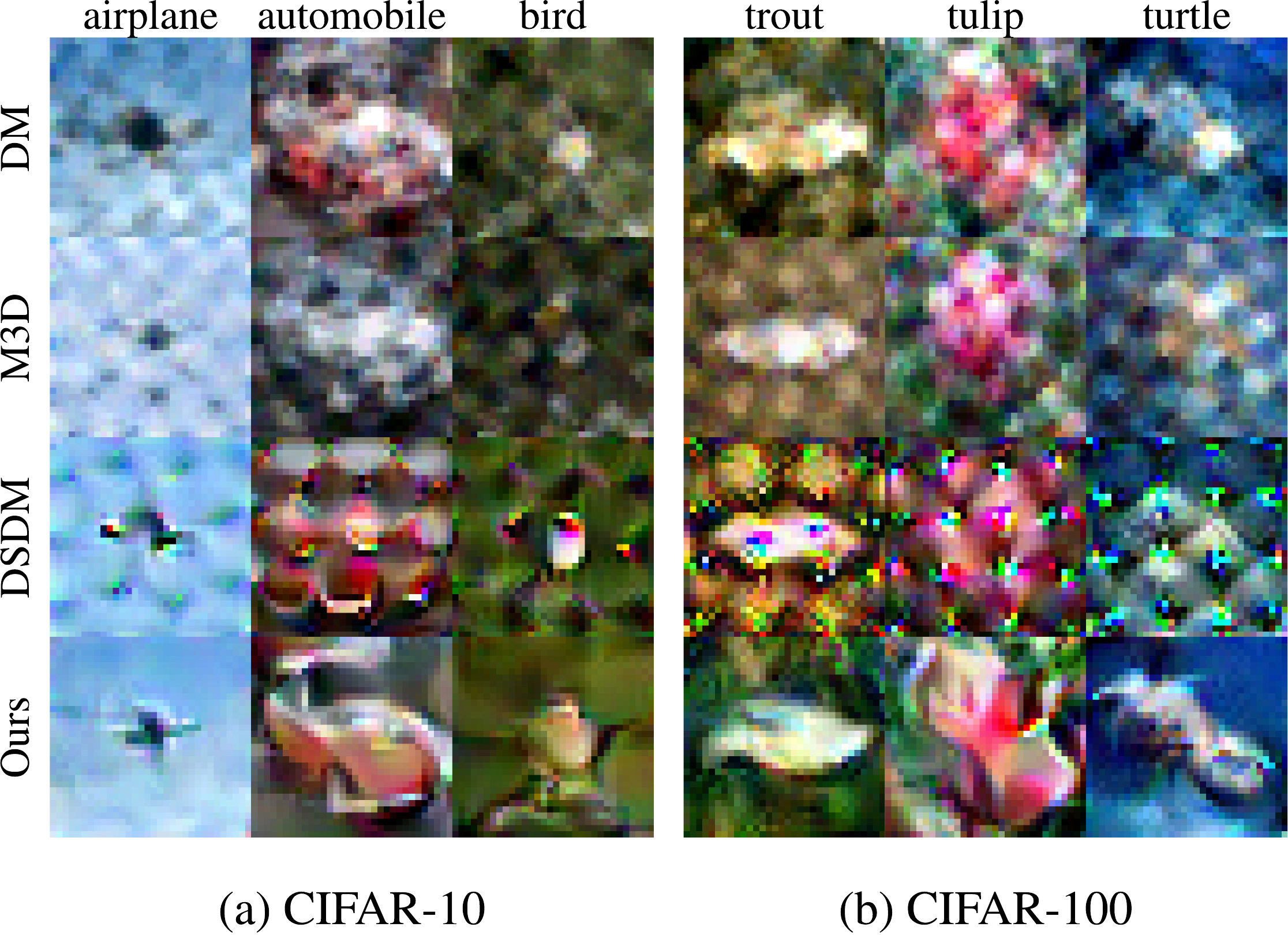}
    \caption{Visualization of synthetic images of CIFAR-10 (partial) and CIFAR-100 (partial) with different method.}
    \label{fig:visulization}
\end{figure}

\subsection{Visualization}
To provide an intuitive qualitative assessment of the distilled images, we visualize the synthetic datasets generated by DM, M3D, DSDM and our method in Figure \ref{fig:visulization}. It can be observed that:
1) The synthetic images produced by our method demonstrate superior structural detail preservation from the original dataset, with substantially reduced noise and enhanced visual clarity.
2) Our synthetic images possess more distinct category-discriminative features, thereby improving inter-class separability.

\section{Conclusion}
In this paper, we revisit the limitations of DM-based dataset distillation, emphasizing their reliance on static representations and the degradation in downstream performance. To address these issues, we propose Trajectory Guided Dataset Distillation (TGDD), a novel approach that dynamically aligns synthetic data with the evolving feature space of the training trajectory. By jointly applying stage-wise distribution matching and distribution constraints, TGDD improves both the diversity and representativeness of synthetic datasets. Extensive experiments on ten benchmarks demonstrate that TGDD delivers superior performance and efficiency, achieving up to a 5.0\% accuracy gain on high-resolution benchmarks. In future work, we will explore how to extend dataset distillation to other modalities and tasks.

\section*{Acknowledgements}
This work was supported in part by the National Science Foundation of China Joint Key Project under Grants U24B20173 and U24B20182, and by the National Natural Science Foundation of China under Grant 62376046, Grant 62536002, Grant 62561160098 and Grant 62402073. It was also supported by the Natural Science Foundation of Chongqing under Grant CSTB2023NSCQ-MSX0341, and by the Science and Technology Research Program of Chongqing Municipal Education Commission under Grant KJQN202300619.

\bibliography{aaai2026}

\end{document}